\documentclass[letterpaper, 10 pt, conference]{ieeeconf}  

\usepackage{booktabs}

\IEEEoverridecommandlockouts                              

\usepackage{graphicx}

\usepackage{etoolbox}
\makeatletter
\patchcmd{\@makecaption}
  {\scshape}
  {}
  {}
  {}
\makeatother

\title{\LARGE \bf
Dense Residual Networks for Gaze Mapping on Indian Roads
}

\author{Chaitanya Kapoor$^{1,*}$, Kshitij Kumar$^{2, *}$, Soumya Vishnoi$^{3, *}$ and Sriram Ramanathan$^{4, *}$ 
\thanks{*Denotes equal contribution}
\thanks{$^{1, 2}$ Dept. of Electrical and Electronics Engineering (BITS Pilani)
        \texttt{\{f20201219, f20190256\} @pilani.bits-pilani.ac.in}}
\thanks{$^{3, 4}$ Dept. of Computer Science and Information Systems (BITS Pilani)
        \texttt{\{f20201512, f20201209\}@pilani.bits-pilani.ac.in}}
}

\begin{document}

\maketitle
\thispagestyle{empty}
\pagestyle{empty}

\begin{abstract}

In the recent past, greater accessibility to powerful computational resources has enabled progress in the field of Deep Learning and Computer Vision to grow by leaps and bounds. This in consequence has lent progress to the domain of Autonomous Driving and Navigation Systems. Most of the present research work has been focused on driving scenarios in the European or American roads. Our paper draws special attention to the Indian driving context. To this effect, we propose a novel architecture, DR-Gaze, which is used to map the driver's gaze onto the road. We compare our results with previous works and state-of-the-art results on the DGAZE dataset. Our code will be made publicly available upon acceptance of our paper.

\end{abstract}

\section{Introduction}
Driving environments can be classified into two major categories $-$ structured and unstructured. The former refers to roads with clearly demarcated boundaries and signs. This also involves well-maintained roads that have a sporadic occurrence of potholes and other forms of damage. 

Unstructured environments on the other hand, as the name suggests, are those that are not bound by a fixed set of rules. This may include the absence of traffic signals, road markings as well as the populace overlooking standard traffic rules. It comes to no surprise that achieving autonomy in such conditions is indeed an elusive task.

Advances in the domain of autonomous driving has seen exponential growth over the past decade, but unfortunately, this is agnostic to a structured context. A significant proportion of such proposed projects like Google's Waymo \cite{Sun_2020_CVPR}, Tesla's \emph{autopilot} \cite{10.1145/3003715.3005465} to name a few have been able to achieve tremendous success in such scenarios.

Several studies have been conducted to demonstrate that inattention and driver fatigue have proven to be a major cause for car crashes \cite{8266142}\cite{Tay2004DRIVERI}. It is of noteworthy importance that these studies have been conducted solely in structured road environments like the USA and Europe. In unstructured driving environments, as is the case in countries like India, it is of paramount importance for drivers to give undivided attention towards the immediate scene in their vicinity. This is a problem that we feel has received a subordinate position by the research community. 

To unriddle this, we propose a novel architecture, \textbf{DR-Gaze} which is used to track the gaze of drivers and map the corresponding coordinates to the immediately visible scene. 

We perform our experiments on the DGAZE dataset \cite{isha2020iros} which is suited for this purpose and is specifically geared towards Indian roads. We also present comparisons with state-of-the-art models on the same task to gauge its efficacy. We hope that our work inspires a new generation of indigenous research gravitating towards the Indian driving context.

\begin{figure}[thpb]
    \centering
    \parbox{3in}{\includegraphics[width=\linewidth]{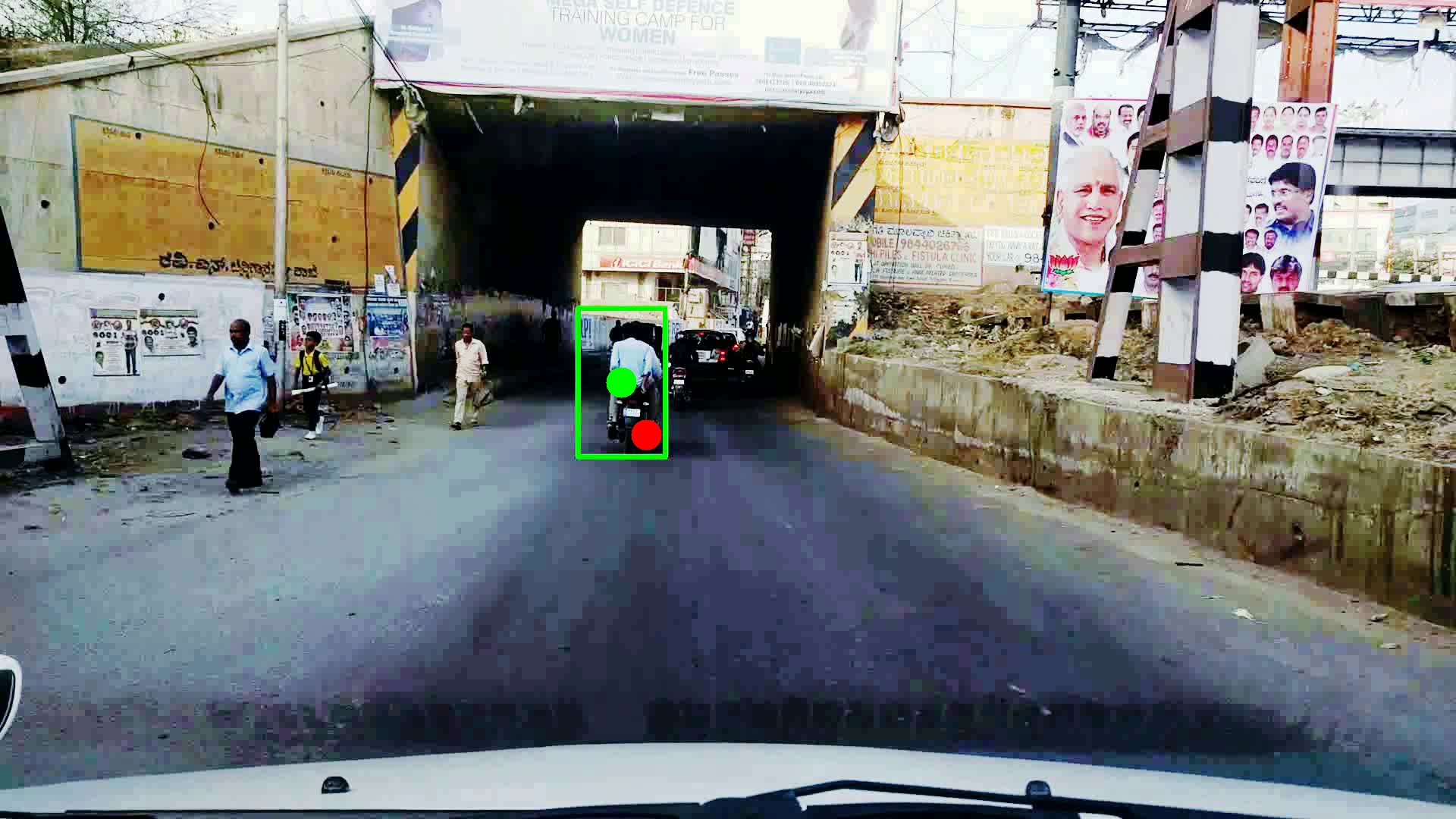}}
    \caption{DR-Gaze result.}
    \label{fig:intro_img}
\end{figure}

\begin{figure*}[t]
    \centering
    \includegraphics[width=0.9\linewidth]{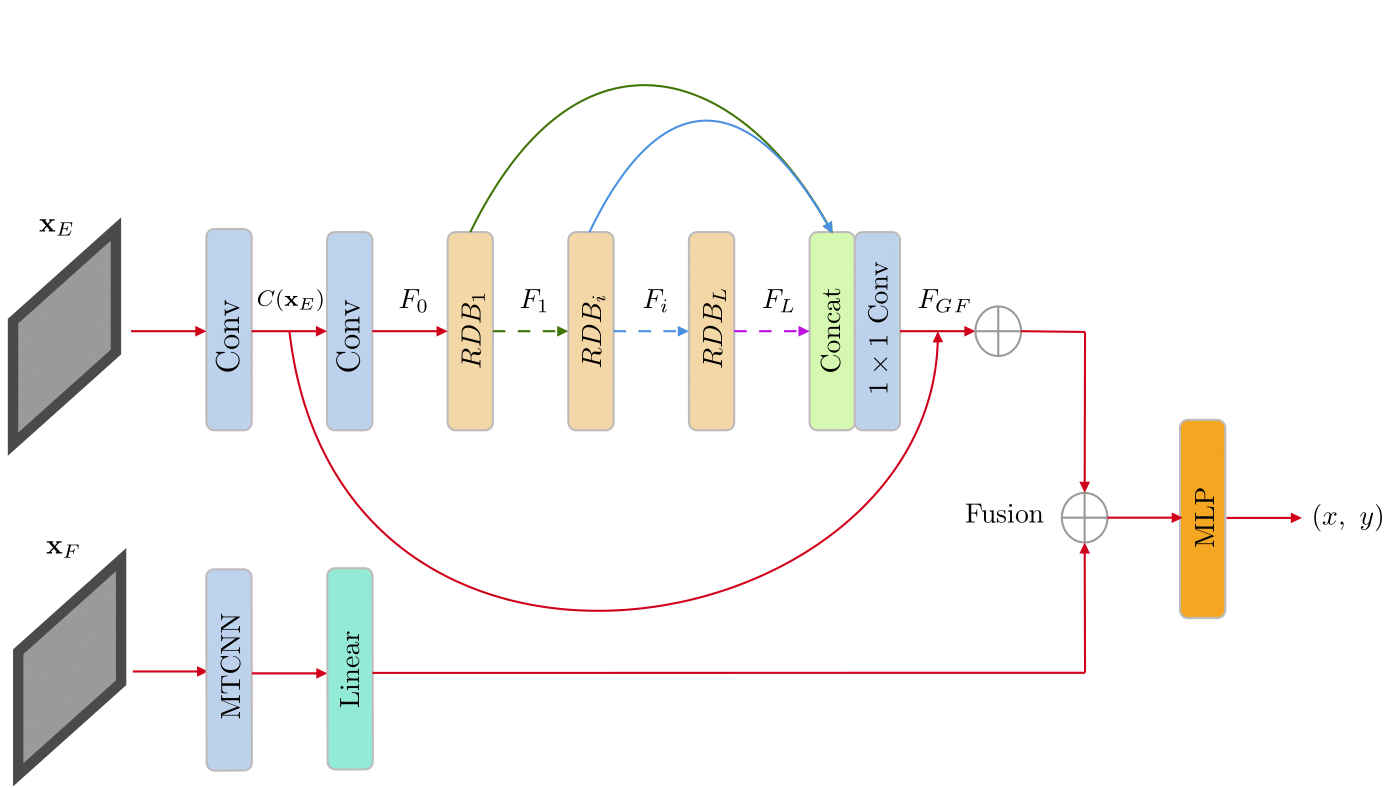}
    \caption{The proposed DR-Gaze architecture. Here, $\mathbf{x}_E$ and $\mathbf{x}_F$ denote the left-eye and facial images respectively.}
    \label{fig:dr-gaze}
\end{figure*}

\section{Related Work}

\textbf{Driver attention monitoring.} It is a known fact that driver distraction is a major cause of road accidents. This has been at the forefront of multiple studies \cite{text_and_drive}, \cite{10.1001/jamapediatrics.2013.1429}. Palazzi et. al. \cite{DBLP:journals/corr/PalazziACSC17} in the DR(eye)VE project have proposed the use of an eye-tracking device and a roof-mounted camera that records the external view onto which the driver's focus is mapped. To capture temporal dependencies, a sequence of frames are given as input to the model and are processed using 3D convolutional layers \cite{3dConv}. There are three fully convolutional branches in the model which predict the FoA (focus of attention) map by processing complementary information from the RGB, motion \cite{optical_flow} and semantic segmentation \cite{semantic_segmentation} domains. \newline Driver attention prediction in accidental scenarios is an extremely critical application. In the DADA (Driver Attention Prediction in Accident Scenarios) model \cite{DBLP:journals/corr/abs-1912-12148} Fang et. al. have utilized mainstream video websites to come up with the DADA-2000 dataset which they use to predict driver attention during accidents. The network for this task has a multi-path 3D encoding (M3DE) architecture, a semantic-guided attentive fusion module (SAF), and a driver attention map decoding module (DAMD). The M3DE consists of two VGG \cite{vgg} like branches that generate spatio-temporal feature maps. The SAF processes these feature maps and combines them in an attentive fusion strategy. The DAMD then produces the final driver attention map.\\


\textbf{Driver gaze mapping} has been interpreted by researchers in a plethora of different ways. Lundgren et. al. \cite{lundgren} proposed a Bayesian filtering approach that used a combination of camera-based driver monitoring systems and filtering techniques to predict the probability of the driver looking at a particular zone. These zones were a broad classification of the regions that drivers could look at from within the car such as the windshield, dashboard or infotainment system. A similar objective has been achieved by Vora et. al. \cite{DBLP:journals/corr/abs-1802-02690} but with the help of CNNs. This method involves pre-processing the input image to obtain a section of it, which predominantly contains the face and in turn serves as the input to the CNNs. The model consists of four different CNN architectures AlexNet \cite{NIPS2012_c399862d}, VGG16 \cite{vgg}, ResNet50 \cite{resnet}, and SqueezeNet \cite{squeeze} which are then evaluated on the Columbia Gaze Dataset \cite{columbia}. An alternative to the problem of driver gaze mapping has been proposed by Chuang et. al  \cite{driver_gaze_smartphone}. Using the front-facing camera of a dashboard-mounted smartphone the driver's face gets captured which is then used to determine the head-pose and obtain a feature vector consisting of the locations of various facial features. The extracted feature vector is then input to a multi-class linear support vector machine that helps learn a binary decision function between every pair of classes, where a class is a portion of the car that the driver is looking at.\\

In this paper, we have taken the task of driver gaze mapping a step further by not only using the driver's image but also taking into account the road image, which is indicative of the scenario that the driver is present in at that particular instant. Both driver and road images of various instances have been used from the DGAZE Dataset for training. This allows the model to input the facial features and head-pose of the driver and map it to a single point on the road image.

\section{Dataset}

The \textbf{DGAZE} dataset consists of two parts, the driver-view videos and the road-view videos. The driver-view and road-view video samples are obtained in a laboratory environment for safety. The road-view videos are common for all the drivers and are annotated before the recording of the driver-view videos. The drivers are then asked to look at these annotated objects in the road-view videos which are projected onto a screen in front of them while the drivers' videos are simultaneously recorded. There are multiple driver-view videos for a single driver and each video corresponds to the driver looking at a particular object. As DGAZE is a video dataset, we have sampled the videos to obtain the first $20$ frames for each driver-view video (the same procedure is followed for the road-view videos) which are then used for training our model.\\
We have split the dataset into the training set that has 11 drivers' data and the validation and test sets that contain 1 driver's data each. With this, we had 22600 images for the training set, 2060 images for the validation set and 2058 images for the test set as opposed to using 98306 image pairs for training, 4779 for validation and 3761 for testing as was done for the I-DGAZE network.

\section{DR-Gaze For Gaze Mapping}
Our network (Fig. \ref{fig:dr-gaze}) is composed of \textbf{2} branches $-$ one for extracting feature vectors from the \textbf{left eye image} and the other for \textbf{extracting facial features and head pose estimation (feature branch)}. We describe each one and its composite blocks in detail below. 

\subsection{Eye branch}
\begin{figure}[thpb]
    \centering
    \parbox{3in}{\includegraphics[width=\linewidth]{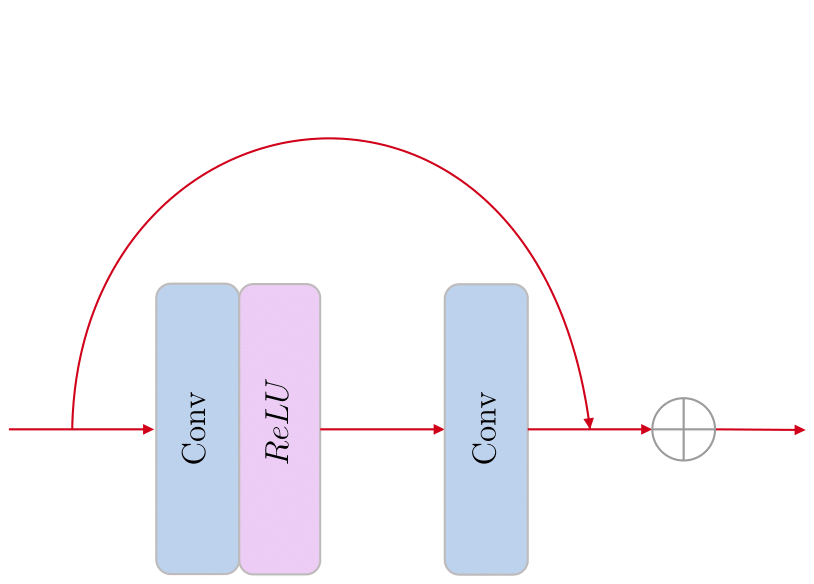}}
    \caption{Residual Block}
    \label{fig:res-block}
\end{figure}

\textbf{Residual block}: He et. al. \cite{resnet} introduced residual mappings $\mathcal{F}(\mathbf{x})$ which are to be learned by the network. The output vector $\mathbf{y}$ of each residual block (Fig. \ref{fig:res-block}) for a said input vector $\mathbf{x}$ is hence
\begin{equation}
    \mathbf{y} = \mathcal{F}(\mathbf{x}, \{W_i\}) + \mathbf{x}
\end{equation}
Here, $W_i$ denotes the weight matrix of the $i^{th}$ layer. The introduction of an identity function improves gradient flow across multiple layers. \\

\begin{figure}[thpb]
    \centering
    \parbox{3in}{\includegraphics[width=\linewidth]{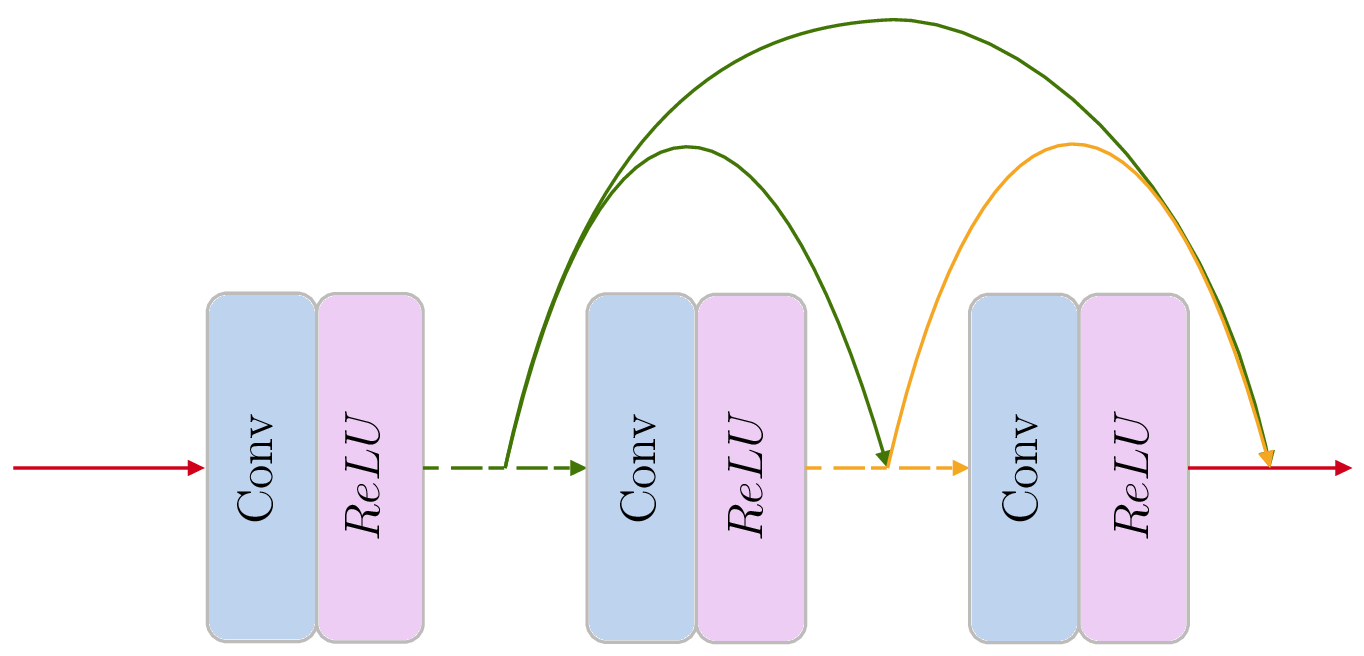}}
    \caption{Dense Block}
    \label{fig:dense-block}
\end{figure}

\textbf{Dense Block}: Huang et. al. \cite{huang2017densely} proposed a different connectivity as opposed to traditional one-to-one connections between subsequent layers for improving information flow. The $\ell^{th}$ layer receives the feature maps ($\mathbf{x}_0, \dots, \mathbf{x}_{\ell - 1}$) of all the preceding $(\ell - 1)$ layers. More concretely, this can be formulated by the following equation
\begin{equation}
    \mathbf{x}_\ell = H_\ell([\mathbf{x}_0,\dots, \mathbf{x}_{\ell - 1}])
\end{equation}
Where $[\mathbf{x}_0,\dots, \mathbf{x}_{\ell - 1}]$ is the concatenation operation. This structure is illustrated in Fig. \ref{fig:dense-block}.\\

\textbf{Residual Dense Block}:
By bringing the best of both worlds \cite{DBLP:journals/corr/abs-1802-08797}, we compose what is called a \textbf{Residual Dense Block} (RDB). \\
Suppose that our network has $N$ residual dense blocks. The output of the $n^{th}$ RDB is given by the recursive relation
\begin{equation}
    F_n = C_i(F_{n-1})
\end{equation}
where $C(\cdot)$ corresponds to a convolution in the $n^{th}$ layer. In our case, the transformation $C$ is characteristically non-linear on account of the introduction of a ReLU non-linearity in each of the individual dense blocks. 

\begin{figure}[thpb]
    \centering
    \parbox{3in}{\includegraphics[width=\linewidth]{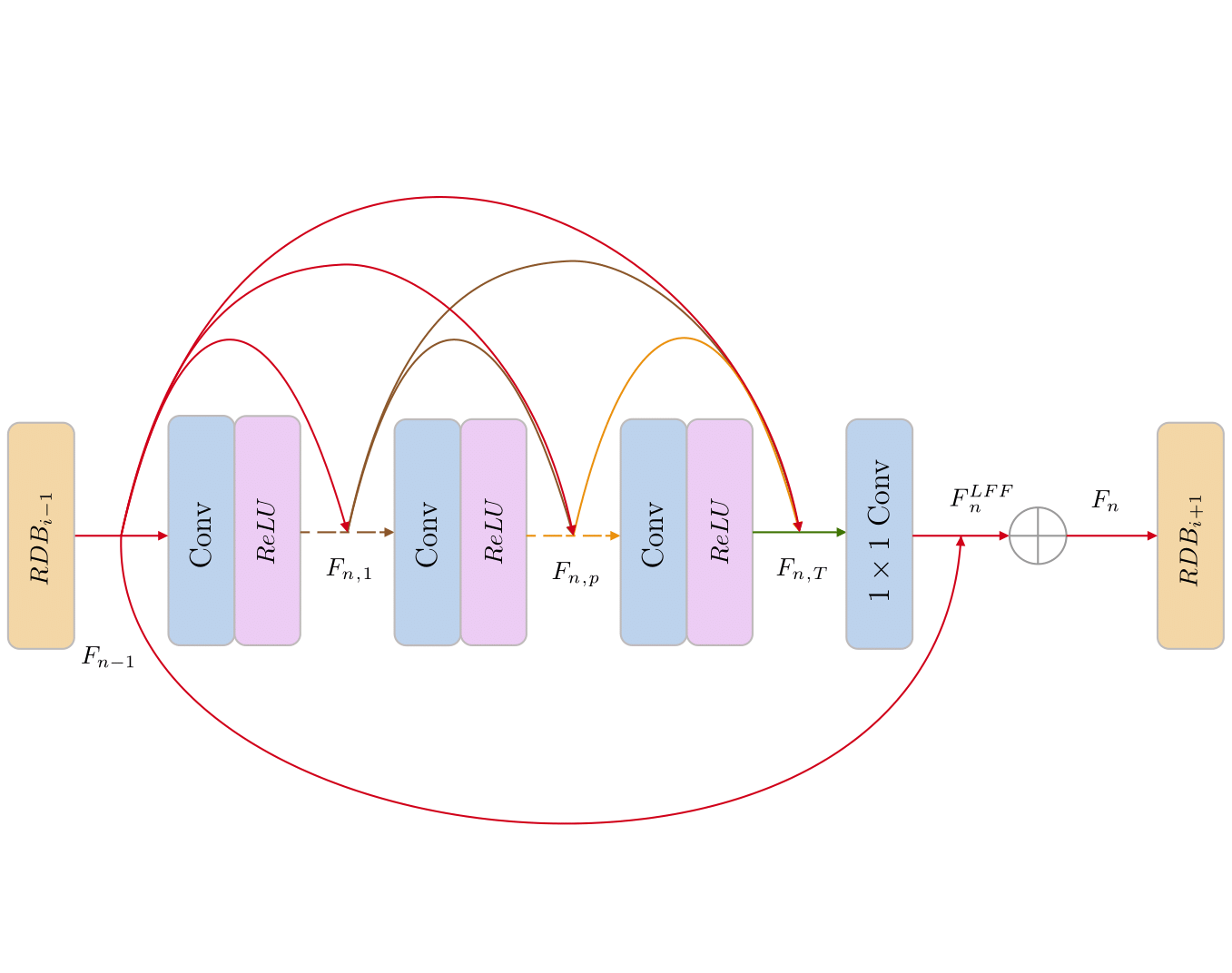}}
    \caption{Residual Dense Block (RDB)}
    \label{fig:rdb}
\end{figure}

A \textbf{Local Feature Fusion} (LFF) is used for adaptive fusion between the current RDB (Fig. \ref{fig:rdb}) and the preceding ones. For controlling the number of parameters resulting after each block, we introduce a $1\times 1$ convolution layer as a bottleneck inspired by ResNet. For the $n^{th}$ RDB, this process can be expressed as (assuming $T$ dense blocks) 
\begin{equation}
    F_{n}^{LFF} = C_n^{LFF}([F_{n-1}, F_{n,1},\dots, F_{n, T}])
\end{equation}

Each of the individual RDBs have local residual learning for improvising information flow amongst layers. For the $n^{th}$ RDB, the resultant output after applying an identity mapping is 
\begin{equation}
    F_n = F_{n-1} + F_n^{LFF}
\end{equation}
Analogous to this, we introduce similar connectivity as in the case of local feature fusion to encompass \textbf{Global Feature Fusion} (GFF). This is used to extract all hierarchical features at a global scale. We fuse all locally fused RDB features into a single feature map which we denote as $F^{GF}$.
\begin{equation}
    F^{GF} = H^{GFF}([F_0^{LFF},F_1^{LFF},\dots,F_n^{LFF}])
\end{equation}
A final global residual mapping is added from the original image with the output of the $N$ RDB blocks.
\begin{equation}
    F = C(\mathbf{x}) + F^{GF}
\end{equation}
where $C(\mathbf{x})$ is the convolved feature map over the original image $\mathbf{x}$. The feature vector so obtained is flattened to a single dimensional vector which we denote by $F_0$ to be passed to subsequent layers.

\subsection{Feature Branch}
The feature branch of the network uses the positions of the driver's facial features as well as the head-pose to get a sense of the direction that the driver is facing. We extract the landmarks for the corners of the eyes, nose, mouth, chin and the coordinates of a bounding box (used as an estimate of facial area). This data is further processed to obtain the values of the pitch, roll and yaw angles of the head \cite{DBLP:journals/corr/ZhangZL016}.
Now we simply concatenate the bounding box coordinates $(x, y, w, h)$, roll, pitch, and yaw angles, coordinates of corners of the eyes and the tip of the nose to obtain a 13-element feature vector. This vector serves as the input to the feature branch of the network. 

\subsection{Feature Fusion}
The feature vectors obtained from MTCNN and Residual Dense Network are simply concatenated with each other to form a 6496 element vector. Formally, the resultant feature vector $F_{0}$ is 
\begin{equation}
    F_0 = [F_1, F_2]
\end{equation}
Where $F_1$ and $F_2$ are the (flattened) feature vectors from MTCNN and RDN respectively, and $[\cdot]$ is the concatenation operation. We then feed this as input to what we call the \textit{``fused branch"} since we essentially fuse two disparate modalities. This is done with the help of a simple 2-layered MLP. We return the gaze coordinates as a 2-tuple consisting of the $(x, y)$ coordinates.
\begin{equation}
    (x, y) = L_2(\sigma(L_1 F_0))
\end{equation}
Here $L_i$ being the $i^{th}$ layer and $\sigma$ is the ReLU non-linearity applied after each linear layer.

\section{Experimental Evaluations And Results}
In this section, we perform a thorough evaluation of DR-Gaze as well as other state-of-the art models on the task of driver gaze prediction. 

We demonstrate that our network outperforms the existing methods whilst being less \emph{data hungry} and requiring lesser training iterations in comparison to others. Our network is able to achieve a per-pixel loss of 180.67 on average. We also demonstrate our results qualitatively to better understand our predictions. We do not perform any calibration on the driver gaze, since our purpose is to allow our network to generalise better.

To train our network, we make use of 13 drivers as opposed to 20. Of these, 11 drivers were reserved for training and 1 driver each for validation and testing respectively. This implies that we have 2 drivers which are completely unseen by our network at the time of evaluation. We have made use of a total of 26,778 image pairs of which 22,600 are used in training, 2060 in validation and 2058 in testing, hence rendering an approximate $80:10:10$ split ratio.

\begin{figure}[thpb]
    \centering
    \parbox{3in}{\includegraphics[width=\linewidth]{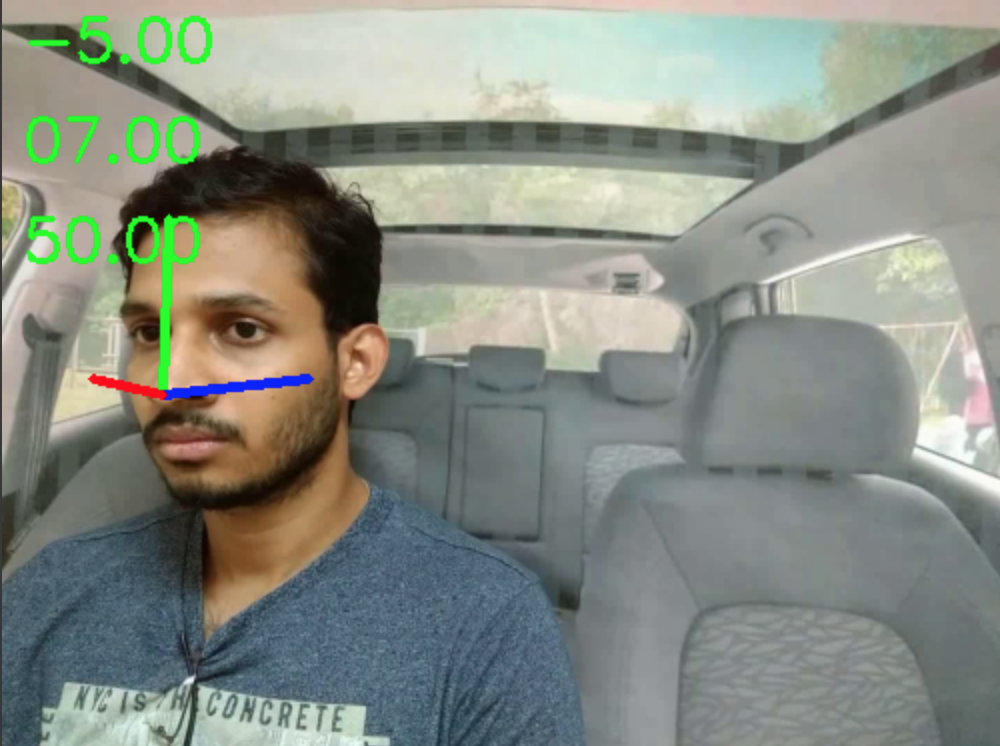}}
    \caption{Head Pose Estimation using MTCNN}
    \label{fig:head-pose}
\end{figure}

\subsection{Implementation Details}
We have made use of two separate networks $-$ one for obtaining feature vectors from the left eye image, and the other for facial feature vectors as detailed in the previous section. We call these network branches the \textbf{Left-Eye} and \textbf{Feature} branches respectively and provide a detailed overview of each branch. 

\textbf{Left-Eye Branch} This branch receives as its input the left eye image of the driver, which has been cropped to a dimension of $36\times 60\times 3$. Our network design borrows ideas heavily from the DenseNet as well as ResNet architecture. We make use of similar nomenclature of parameters whilst stating them here. For evaluating our network, we set the number of channels $(c=3)$, number of features $(f=32)$, number of blocks $(b=32)$, growth rate $(k=4)$ and the number of layers $(l=4)$.

\textbf{Feature Branch} We estimate the head pose of the driver by fine-tuning MTCNN \cite{DBLP:journals/corr/ZhangZL016}. This was implemented using OpenCV and Dlib \cite{headposegithub}. The 13-element feature vector obtained from this branch is passed through a linear layer and transformed to a 16-element vector which we concatenate with the feature maps from the left-eye branch.

\textbf{MLP} Our implementation makes use of a 2-layered neural network. The primary layer and penultimate layers have 6496 and 500 input features respectively. The final layer returns a 2-tuple signifying the $(x,y)$ gaze coordinates. Note that as stated before, a ReLU non-linearity is applied between subsequent linear layers.

\textbf{Training} We make use of the Adam optimizer \cite{kingma2017adam} where we set the value of learning rate $\alpha = 1\mathrm{e}{-5}$ and $\epsilon = 1\mathrm{e}{-5}$. We also use a MultiStepLR scheduler where we initialize the rate decay $\gamma = 0.1$ and the milestone epoch indices in the range $(40-55)$. 

To ensure faster convergence, we made use of Xavier initialization \cite{Glorot10understandingthe}. Further, we used a batch size of $32$ to run our experiments. We have also normalized each driver's image using its mean and standard deviation.

\begin{figure*}[t]
    \centering
    \setlength\tabcolsep{1.0pt}
    \begin{tabular}{ccc}
        \includegraphics[width=0.32\textwidth]{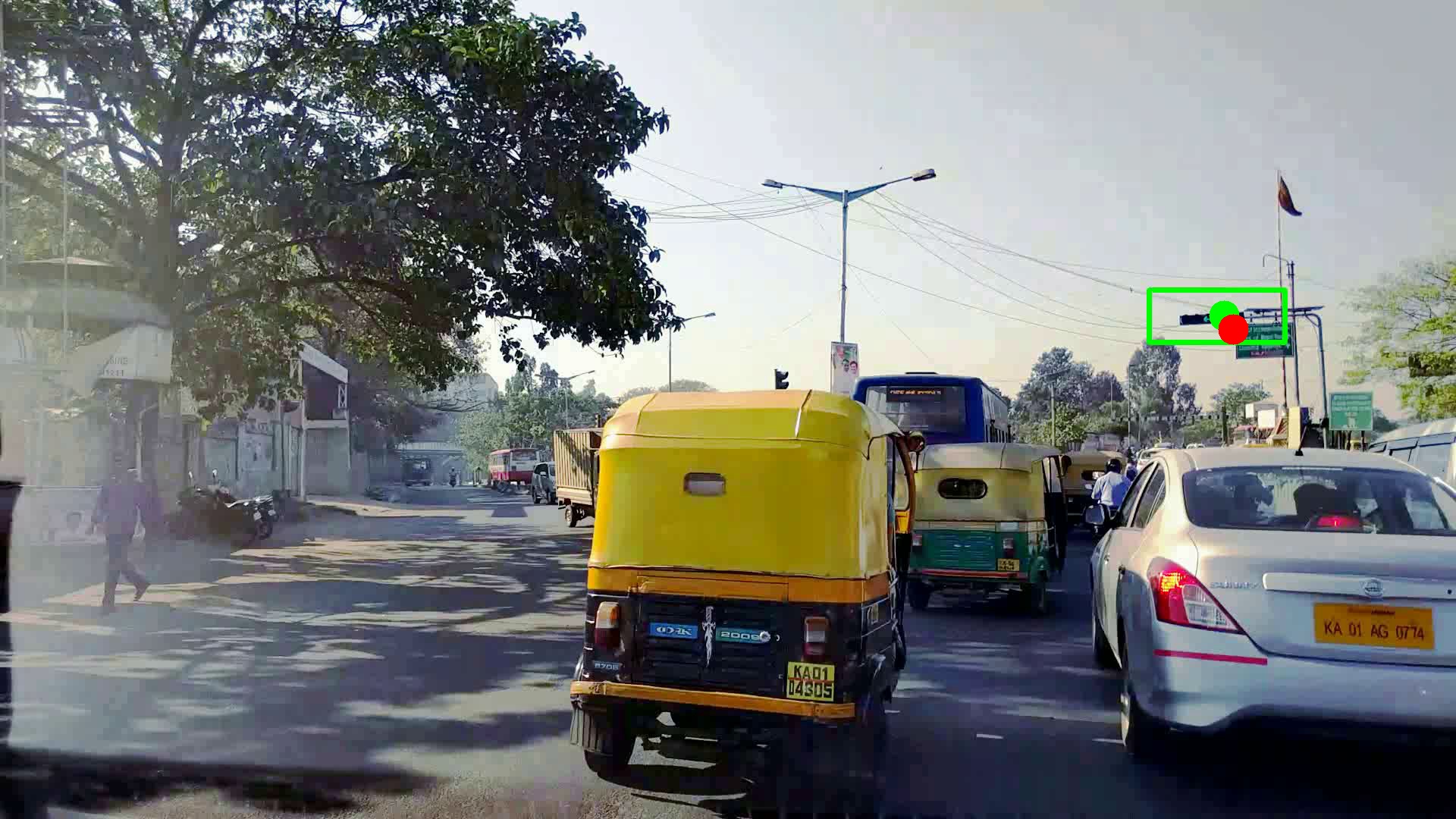} & \includegraphics[width=0.32\textwidth]{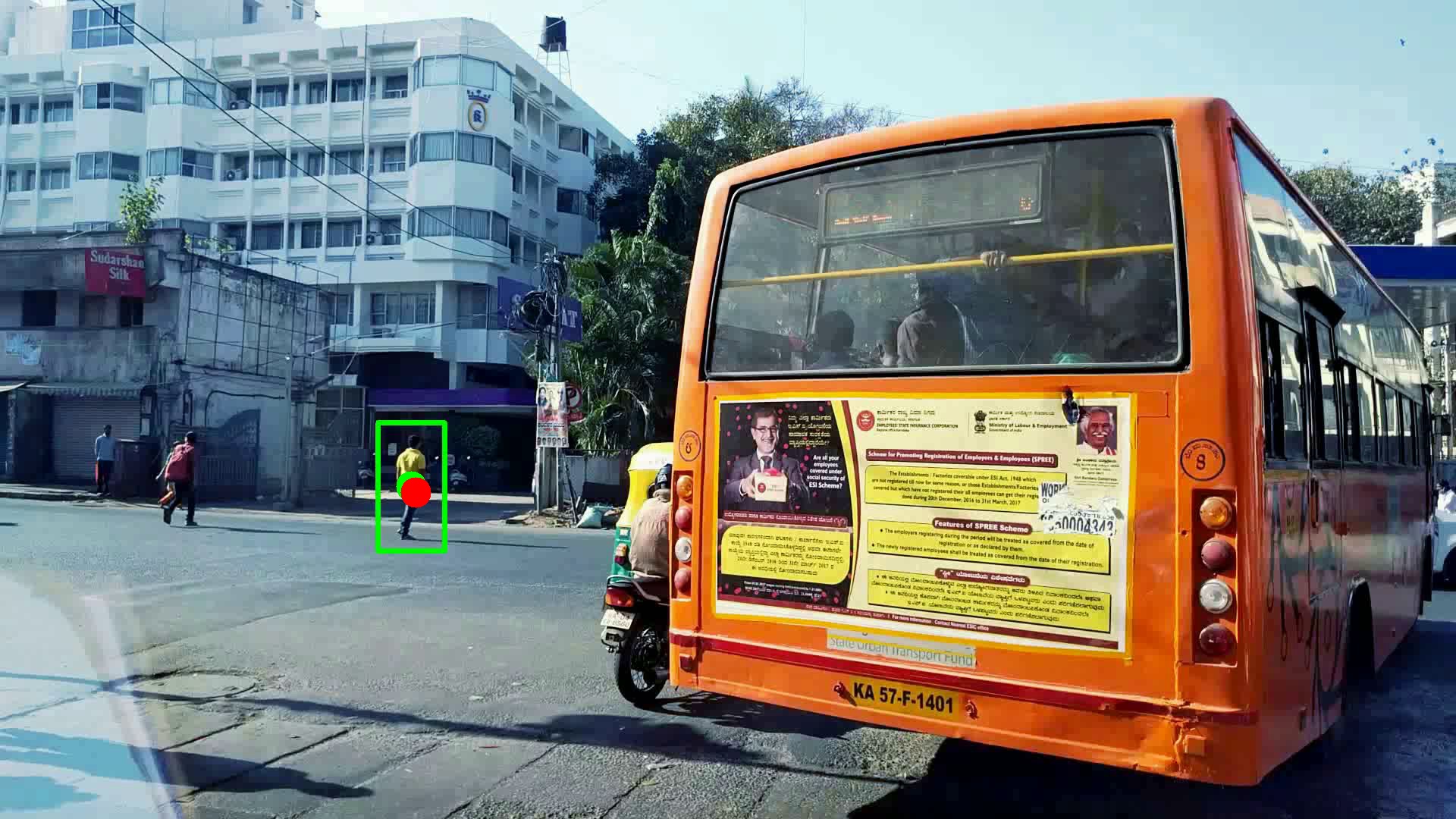} & \includegraphics[width=0.32\textwidth]{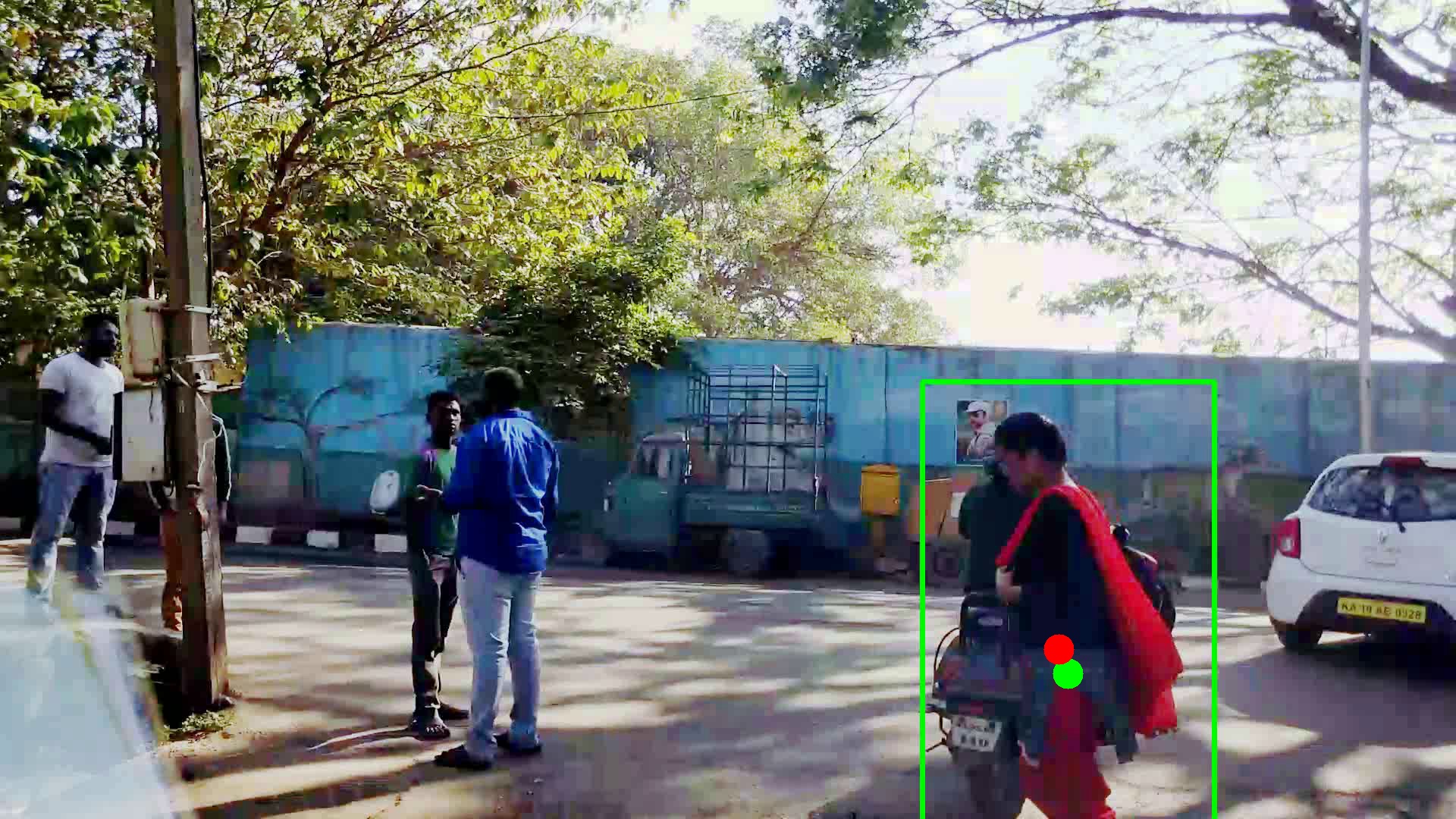}\\
        
        \includegraphics[width=0.32\textwidth]{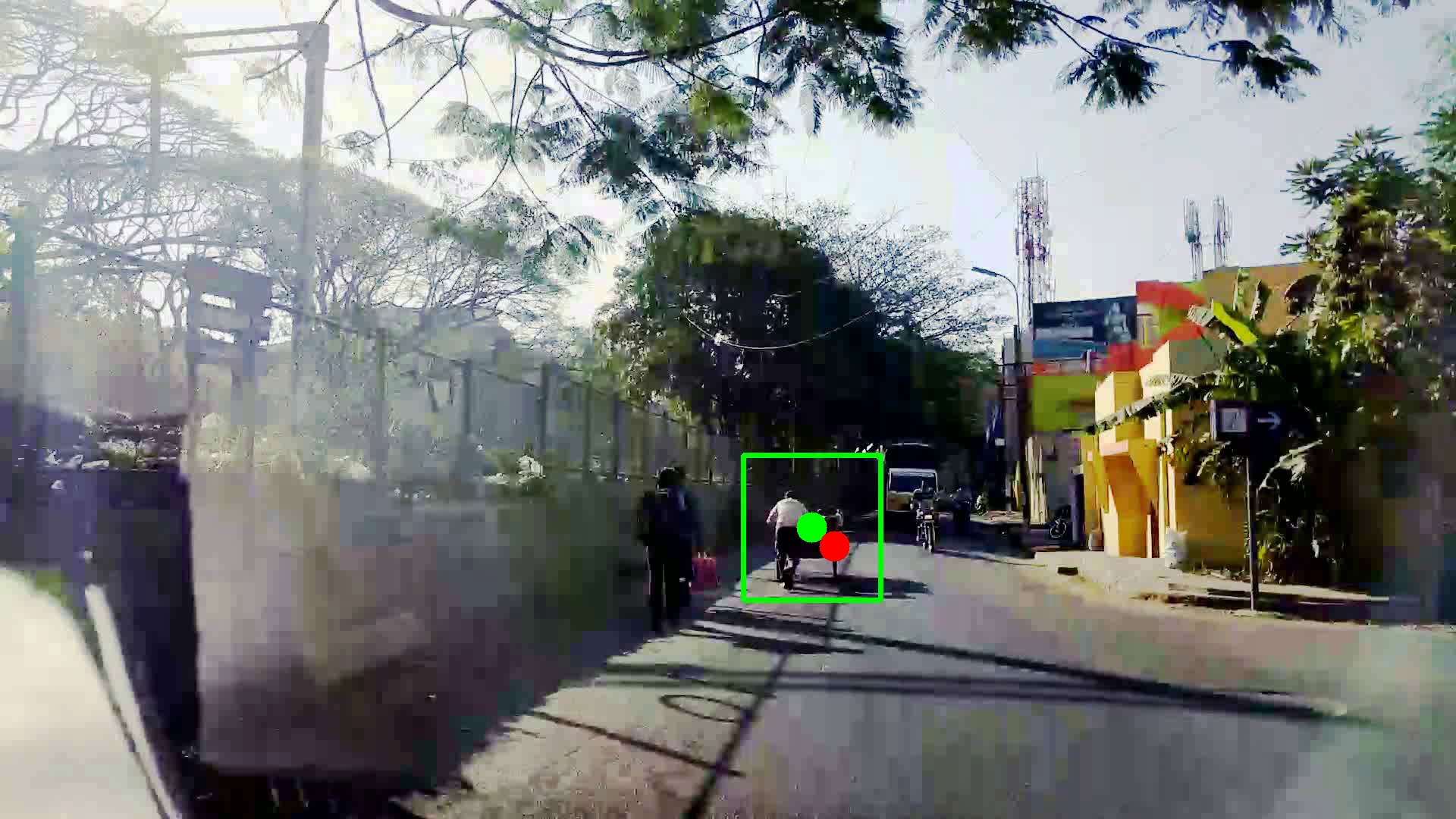} & \includegraphics[width=0.32\textwidth]{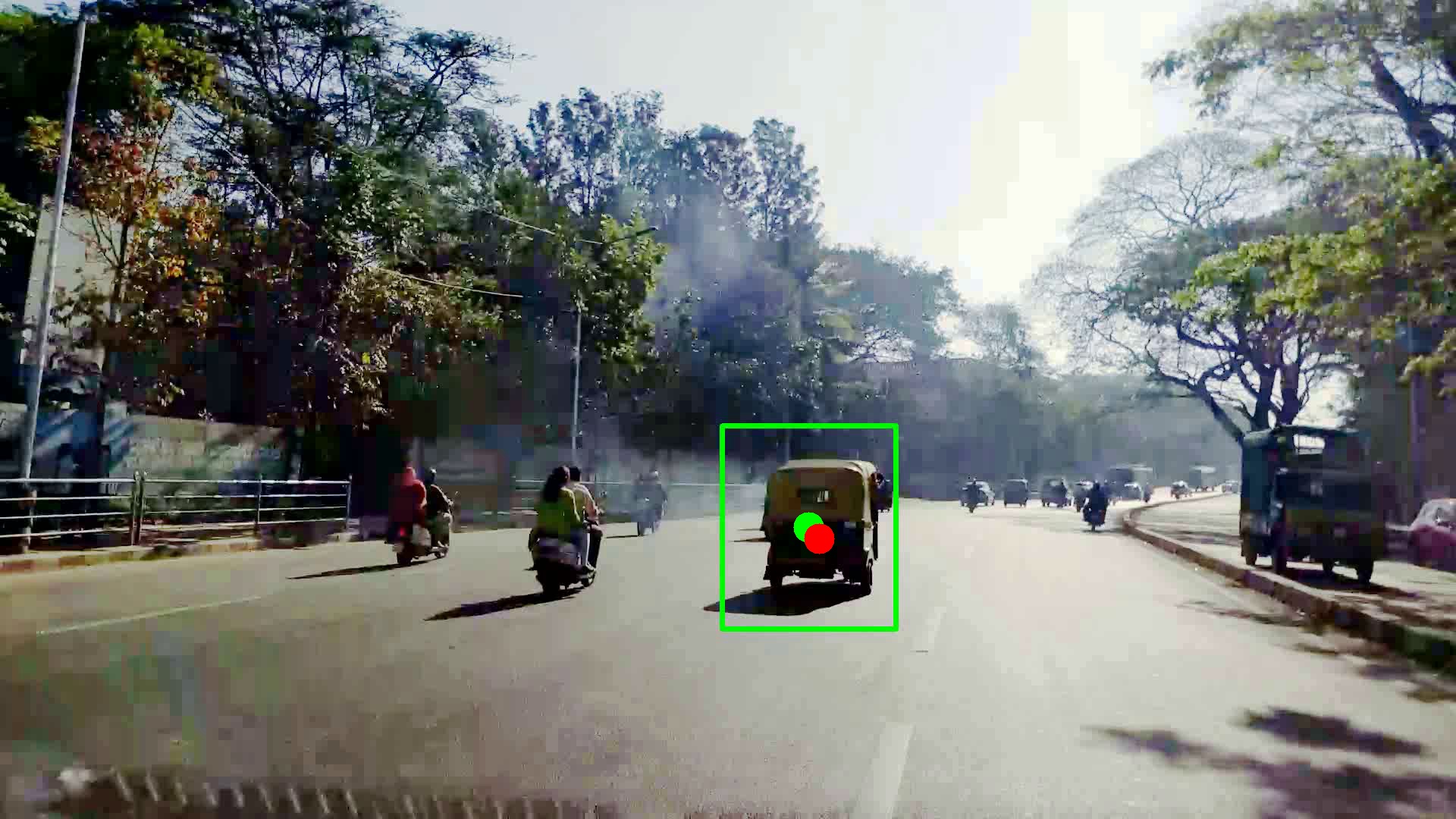} & \includegraphics[width=0.32\textwidth]{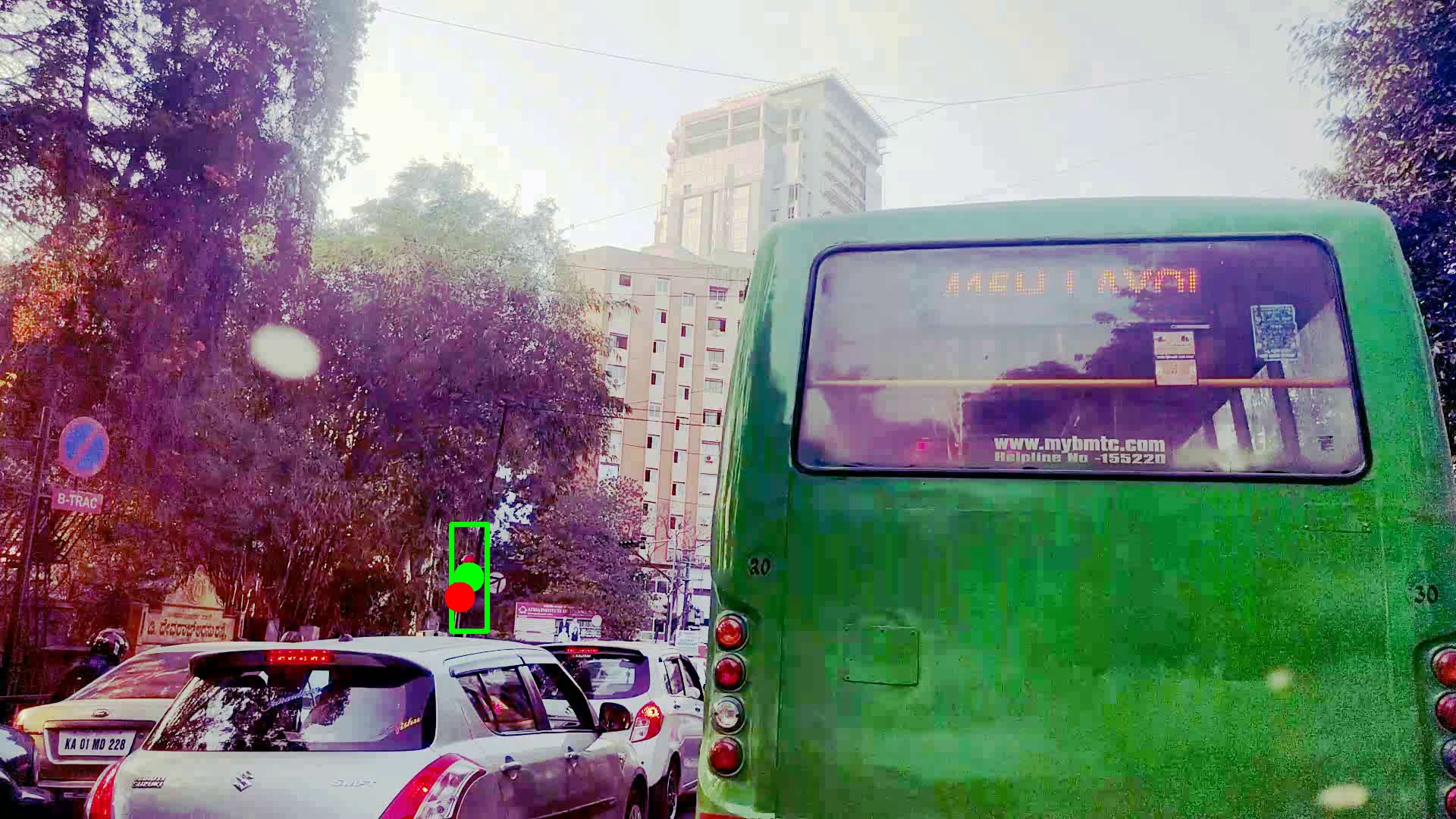}\\
        
        \includegraphics[width=0.32\textwidth]{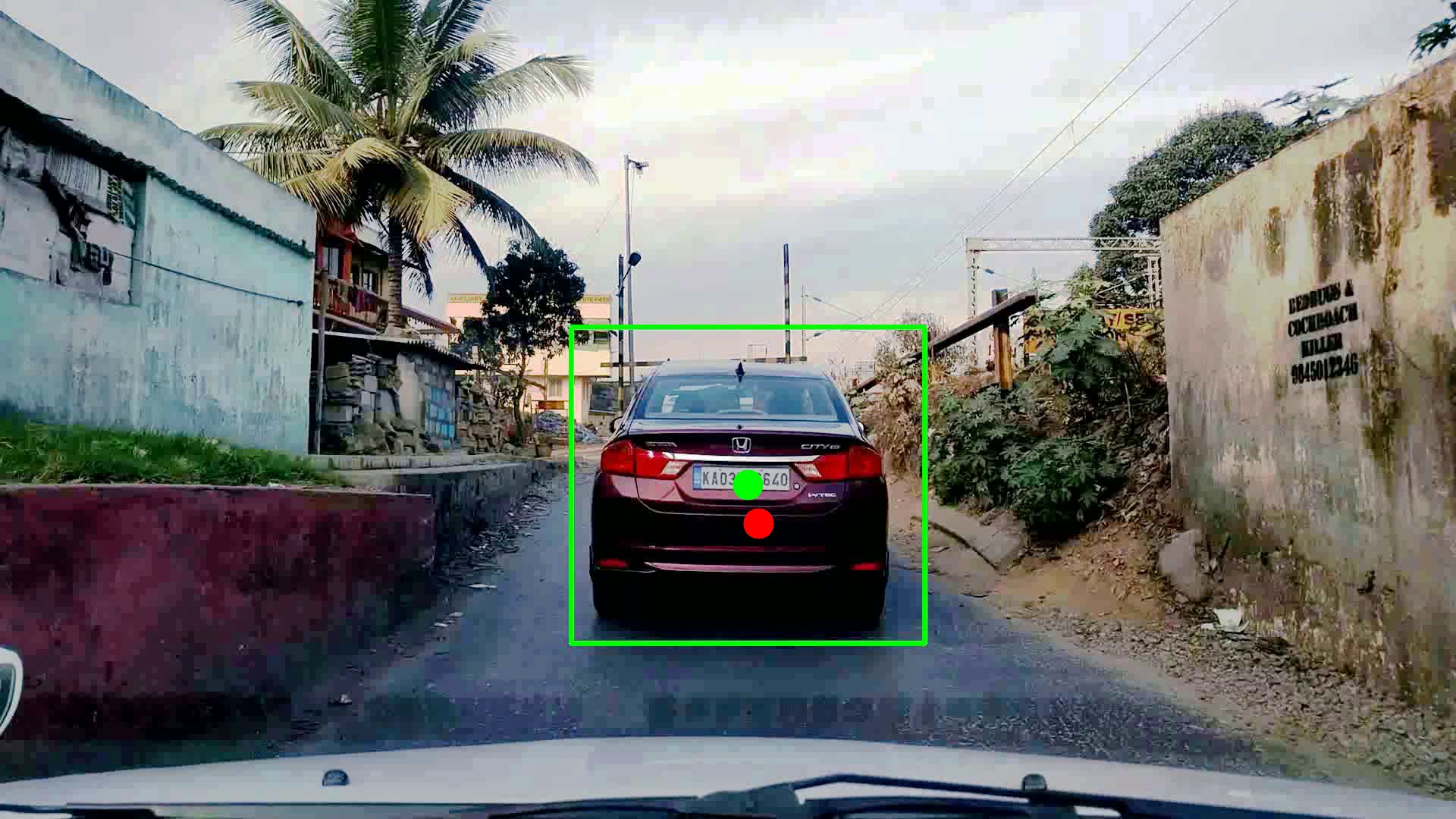} & \includegraphics[width=0.32\textwidth]{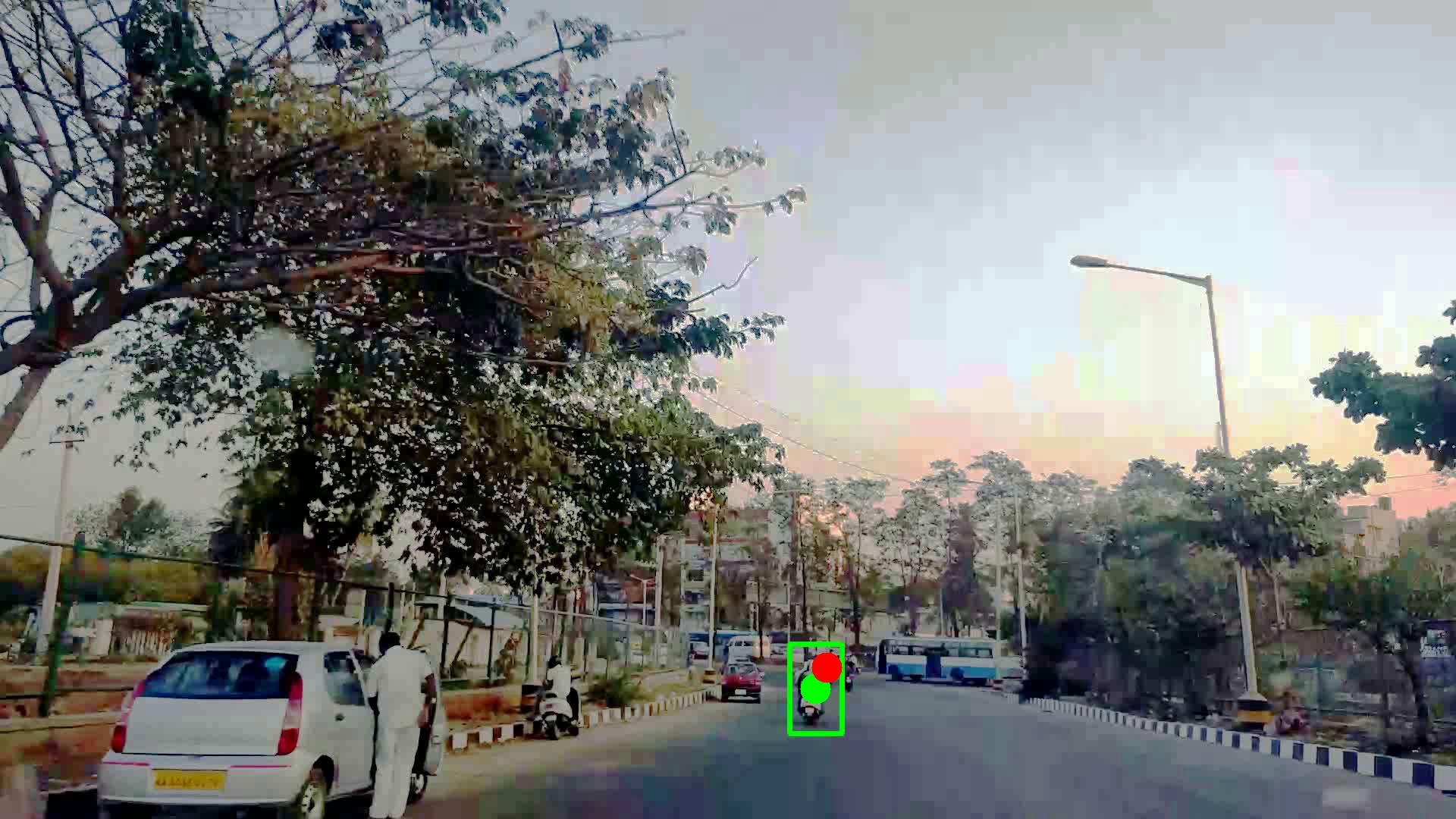} & \includegraphics[width=0.32\textwidth]{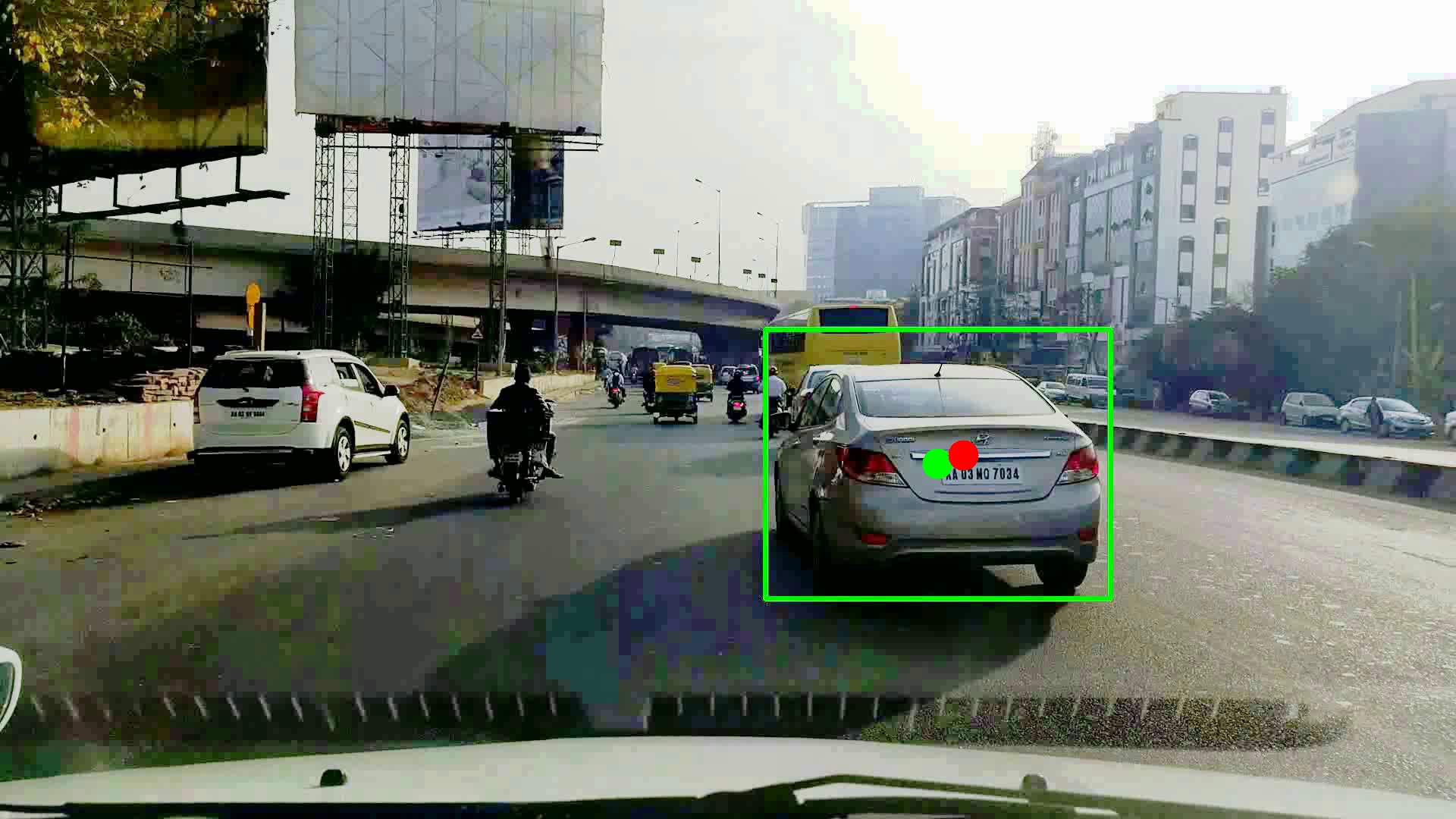}\\
    \end{tabular}

    \caption{Gaze predictions of DR-GAZE. The green point corresponds to the ground truth prediction, and the red points denote the predicted gaze coordinate.}
    \label{fig:results-collage}
\end{figure*}

\subsection{Evaluation Metrics}

To ensure a fair comparison with existing state-of-the-art networks, we evaluate our model on the standard absolute $L_1$ loss which is given by 
\begin{equation}
    \mathcal{L} = \frac{1}{n}\sum_{i=1}^{n}|Y_i - \hat{Y}_i|
\end{equation}
This ensures and yields a per-pixel loss on our gaze prediction with respect to the ground truth coordinates. It must be noted that the resolution of the images is $1920\times 1080$, and hence the per-pixel loss must be interpreted at this scale. We forgo the calibration step since our goal is centred on building a relatively general network. Calibration was performed to make the network fine-tuned for each driver. Thus, unlike I-DGAZE, our network provides generalized results and is driver-agnostic.

\subsection{Results}
We provide a quantitative comparison of our model with other state-of-the-art methods in Table \ref{table:1}. 

\setlength{\tabcolsep}{0.75em}
{\renewcommand{\arraystretch}{1.4}
\begin{table}[h!]
\centering
\begin{tabular}{cccc} \toprule
    {Method} & {Train error} & {Val. error} & {Test error}  \\ \midrule
    Turker Gaze & $171.300$ & $176.37$ & $190.717$\\
    MPII Gaze & $144.32$ & $229.0$ & $189.63$\\
    iTracker & $140.1$ & $205.65$ & $190.5$\\
    I-DGAZE & $133.34$ & $204.77$ & $186.89$\\ 
    DR-Gaze & $\mathbf{88.735}$ & $\mathbf{180.67}$ & $\mathbf{184.689}$\\
    \bottomrule
\end{tabular}
\caption{Experimental evaluations and comparison of our network with various state-of-the-art methods.}
\label{table:1}
\end{table}
}

All the methods listed vary significantly from one another in their approach to computing driver gaze points. Turker Gaze \cite{DBLP:journals/corr/GirshickDDM13} method makes use of ridge regression for gaze-point estimation. For this network, pixel-level facial features serve as the input. We contrast and compare our results with MPII Gaze \cite{7299081} which employs a multimodal convolutional neural network for appearance-based gaze estimation in the wild. Lastly, iTracker \cite{DBLP:journals/corr/KrafkaKKKBMT16} is a convolutional neural network for eye tracking which has claimed several state-of-the-art results.

All our results tabulated are \textbf{without} performing any form of calibration. Thus, we observe that DR-Gaze delivers better performance across all categories and yields better generalization power, which we attribute to the various stages of feature fusion (local and global) as detailed in section IV.

\section{CONCLUSIONS}

In this paper, we proposed a novel architecture to tackle the problem of driver gaze mapping. Our model \textbf{DR-Gaze} which has been trained on the DGAZE dataset,  produced better results than state-of-the-art models all the while being computationally less expensive as well as requiring a lesser amount of training data. The model is a late-fusion based two-branch network that takes the left-eye image and a facial feature vector as input for the branches respectively. These are then processed to output the coordinates of the point that the driver is looking at.
While existing models predict regions within the car that the driver looks at, our model is more specific and can map the driver's gaze to a single point on the road image with high accuracy.\newline
We believe that our work will inspire researchers to come up with innovative solutions in the areas of gaze mapping, accident prevention, driver attention prediction, etc., thereby helping us in moving a step closer to achieving the goal of a fully-autonomous driving system.


\section*{ACKNOWLEDGMENT}

This research was supported by Sally Robotics (autonomous navigation research group) and the Center for Robotics and Intelligent Systems (CRIS) Lab at BITS Pilani. We wish to express our gratitude to BITS Pilani for their constant support during the course of research.

\bibliographystyle{IEEEtran} 
\bibliography{IEEEexample}

\end{document}